\pdfoutput=1

\documentclass[10pt,english,letterpaper, journal, twoside]{IEEEtran}
\usepackage[T1]{fontenc}
\usepackage[latin1]{inputenc}
\usepackage{xcolor}
\usepackage{float}
\usepackage{units}
\usepackage{mathtools}
\usepackage{amsmath}
\usepackage{amsthm}
\usepackage{amssymb}
\usepackage{cancel}
\usepackage{mathdots}
\usepackage{stmaryrd}
\usepackage{stackrel}
\usepackage{graphicx}
\PassOptionsToPackage{version=3}{mhchem}
\usepackage{mhchem}
\usepackage{esint}

\makeatletter

\let\oldforeign@language\foreign@language
\DeclareRobustCommand{\foreign@language}[1]{%
  \lowercase{\oldforeign@language{#1}}}
\theoremstyle{plain}
\newtheorem{thm}{\protect\theoremname}
\theoremstyle{plain}
\newtheorem{lem}[thm]{\protect\lemmaname}

\usepackage{color}
\usepackage{xcolor}
\usepackage{graphicx}
\usepackage{import}
\graphicspath{{images/}}

\IEEEoverridecommandlockouts

\makeatother

\usepackage{babel}
\providecommand{\lemmaname}{Lemma}
\providecommand{\theoremname}{Theorem}

\begin{document}
\global\long\def\dq#1{\underline{\boldsymbol{#1}}}%

\global\long\def\quat#1{\boldsymbol{#1}}%

\global\long\def\mymatrix#1{\boldsymbol{#1}}%

\global\long\def\myvec#1{\boldsymbol{#1}}%

\global\long\def\mapvec#1{\boldsymbol{#1}}%

\global\long\def\dualvector#1{\underline{\boldsymbol{#1}}}%

\global\long\def\dual{\varepsilon}%

\global\long\def\dotproduct#1{\langle#1\rangle}%

\global\long\def\norm#1{\left\Vert #1\right\Vert }%

\global\long\def\mydual#1{\underline{#1}}%

\global\long\def\getp#1{\operatorname{\mathcal{P}}\left(#1\right)}%

\global\long\def\getd#1{\operatorname{\mathcal{D}}\left(#1\right)}%

\global\long\def\hamilton#1#2{\overset{#1}{\operatorname{\mymatrix H}}\left(#2\right)}%

\global\long\def\hamiquat#1#2{\overset{#1}{\operatorname{\mymatrix H}}_{4}\left(#2\right)}%

\global\long\def\hami#1{\overset{#1}{\operatorname{\mymatrix H}}}%

\global\long\def\tplus{\dq{{\cal T}}}%

\global\long\def\swap#1{\text{swap}\{#1\}}%

\global\long\def\imi{\hat{\imath}}%

\global\long\def\imj{\hat{\jmath}}%

\global\long\def\imk{\hat{k}}%

\global\long\def\real#1{\operatorname{\mathrm{Re}}\left(#1\right)}%

\global\long\def\imag#1{\operatorname{\mathrm{Im}}\left(#1\right)}%

\global\long\def\imvec{\boldsymbol{\imath}}%

\global\long\def\vector{\operatorname{vec}}%

\global\long\def\mathpzc#1{\fontmathpzc{#1}}%

\global\long\def\cost#1#2{\underset{\text{#2}}{\operatorname{\text{cost}}}\left(\ensuremath{#1}\right)}%

\global\long\def\diag#1{\operatorname{diag}\left(#1\right)}%

\global\long\def\frame#1{\mathcal{F}_{#1}}%

\global\long\def\ad#1#2{\text{Ad}\left(#1\right)#2}%

\global\long\def\norm#1{\left\Vert #1\right\Vert }%

\global\long\def\minim#1#2{ \begin{aligned} &  \underset{#1}{\min}  &   &  #2 \end{aligned}
 }%

\global\long\def\abs#1{\left|#1\right|}%

\global\long\def\minimone#1#2#3{ \begin{aligned} &  \underset{#1}{\min}  &   &  #2 \\
  &  \text{subject to}  &   &  #3 
\end{aligned}
 }%

\global\long\def\minimtwo#1#2#3#4{ \begin{aligned} &  \underset{#1}{\min}  &   &  #2 \\
  &  \text{subject to}  &   &  #3 \\
  &   &   &  #4 
\end{aligned}
 }%

\global\long\def\minimtwom#1#2#3#4{ \begin{aligned} &  \underset{#1}{\min}  &   &  #2 \\
  &  \text{subject to}  &   &  #3 \\
  &   &   &  #4 
\end{aligned}
 }%

\global\long\def\minimthree#1#2#3#4#5{ \begin{aligned} &  \underset{#1}{\min}  &   &  #2 \\
  &  \text{subject to}  &   &  #3 \\
  &   &   &  #4 \\
  &   &   &  #5 
\end{aligned}
 }%

\global\long\def\argmin#1#2#3#4{ \begin{aligned}#4  &  \underset{#1}{\arg\min}  &   &  #2\\
  &  \text{subject to}  &   &  #3 
\end{aligned}
 }%

\title{Whole-Body Control with (Self) Collision Avoidance Using Vector Field
Inequalities}
\author{Juan José Quiroz-Omaña and Bruno Vilhena Adorno, \IEEEmembership{Senior~Member,~IEEE}\thanks{Manuscript
received: February 24, 2019; Revised: May 30, 2019; Accepted: June
28, 2019.}\thanks{This paper was recommended for publication by
Editor Paolo Rocco upon evaluation of the Associate Editor and Reviewers
comments.}\thanks{This work was supported by the Brazilian agencies
CAPES, CNPq (424011/2016-6 and 303901/2018-7), FAPEMIG, and by the
INCT (National Institute of Science and Technology) under the grant
CNPq (Brazilian National Research Council) 465755/2014-3. }\thanks{J.J.
Quiroz Omaña is with the Graduate Program in Electrical Engineering
- Federal University of Minas Gerais (UFMG) - Av. Antônio Carlos 6627,
31270-901, Belo Horizonte-MG, Brazil. Email: \texttt{juanjqo@ufmg.br}.
B.V. Adorno is with the Department of Electrical Engineering - Federal
University of Minas Gerais (UFMG) - Av. Antônio Carlos 6627, 31270-901,
Belo Horizonte-MG, Brazil. Email: \texttt{adorno@ufmg.br}.}\thanks{Digital
Object Identifier (DOI): see top of this page.}}
\markboth{IEEE ROBOTICS AND AUTOMATION LETTERS. PREPRINT VERSION. ACCEPTED JULY,
2019}{Quiroz-Omaña \MakeLowercase{\emph{et al.}}: Whole-Body Control with
(Self) Collision Avoidance Using Vector Field Inequalities}
\maketitle
\begin{abstract}
This work uses vector field inequalities (VFI) to prevent robot self-collisions
and collisions with the workspace. Differently from previous approaches,
the method is suitable for both velocity and torque-actuated robots.
We propose a new distance function and its corresponding Jacobian
in order to generate a VFI to limit the angle between two Plücker
lines. This new VFI is used to prevent both undesired end-effector
orientations and violation of joints limits. The proposed method is
evaluated in a realistic simulation and on a real humanoid robot,
showing that all constraints are respected while the robot performs
a manipulation task.
\end{abstract}

\section{Introduction}

Collision avoidance has been addressed either in off-line or on-line
approaches. The former is based on motion planning, where probabilistic
methods have been widely used \cite{May2011,Burget2016}. These strategies
are usually applied in the configuration space and are computationally
expensive, free of local minima and used in structured scenarios \cite{Moll2015}.
The latter is based on reactive methods and usually require less computation
time; therefore, they can be used in real-time feedback control and
are suitable for applications within unstructured workspaces.

Reactive methods are, in general, based on minimization problems \cite{Laumond2015},
which exploit the robot redundancy by selecting admissible control
inputs based on a specified criterion. When the robot is commanded
by joint velocity inputs and operates under relatively low velocities
and accelerations, its behavior is appropriately described by the
kinematic model and the minimization can be performed in the joint
velocities. In that case, since the control law is based entirely
on the kinematic model, it is not affected by uncertainties in the
inertial parameters (e.g., mass and moment of inertia). On the other
hand, if the robot is commanded by torque inputs, the minimization
is performed in the joint torques, which usually requires the robot
dynamic model. Both methods are widely used to perform whole-body
control with reactive behavior.

Whole-body control strategies with collision avoidance usually have
been handled by using the task priority framework, where the overall
task is divided into subtasks with different priorities. For instance,
distance functions with continuous gradients between convex hulls
(or between simple geometrical primitives such as spheres and cylinders)
that represent the body parts are used and the lower-priority collision-avoidance
tasks are satisfied in the null space of higher-priority ones \cite{Stasse2008,Schwienbacher2011}.
Those secondary tasks are fulfilled as long as they are not in conflict
with the higher-priority ones. Therefore, they do not prevent collisions
when in conflict with the primary task. One way to circumvent this
problem is to place the collision avoidance as the higher priority
task \cite{Sentis2004}, at the expense of not guaranteeing the fulfillment
of the main task, such as reaching targets with the end-effector.
This can be addressed by using a dynamic task prioritization \cite{Mansard2009},
where the control law is blended continuously between the collision-avoidance
task and the end-effector pose control task as a function of the collision
distance \cite{Sugiura2007}. Dietrich et al. \cite{Dietrich2012}
propose a torque-based self-collision avoidance also using dynamic
prioritization, where the transitions are designed to be continuous
and comply with the robot's physical constraints, such as the limits
on joint torque derivatives. However, changing priorities to enforce
inequality constraints has an exponential cost in the number of inequalities
\cite{Kanoun2011}. Alternatively, non-hierarchical formulations based
on weighted least-square solutions are used to solve the problem of
constrained closed-loop kinematics in the context of self-collision
avoidance \cite{Patel2005}. However, the non-hiearchical approach
requires an appropriate weighting matrix that results in a collision-free
motion, which may not work in general, specially for high-speed motions
\cite{Dariush2010}. That can be solved by combining the weighting
matrix with the virtual surface method, which redirects the collision
points along a virtual surface surrounding the robot links at the
expense of potentially disturbing the trajectory tracking when the
distance between collidable parts is smaller than a critical value.

Other strategies address the collision-free motion generation problem
explicitly by using mathematical programming, which allows dealing
with inequality constraints directly in the optimization formulation,
providing an efficient and elegant solution, where all constraints
are clearly separated from the main task. Analytical solutions, however,
usually do not exist and numeric solvers must be used \cite{kim2013,Escande2014,Fraisse2015a}.

Marinho et al. \cite{Marinho2018} propose active constraints based
on VFI to deal with collision avoidance in surgical applications.
Both robot and obstacles are modeled by using geometric primitives---such
as points, planes, and Plücker lines---, and distance functions with
their respective Jacobian matrices are computed from these geometric
primitives.

Our work extends the VFI method, which was first proposed using first
order kinematics \cite{Marinho2018}, to use second order kinematics.
This enables applications that use the robot dynamics by means of
the relationship between joint torques and joint accelerations in
the Euler-Lagrange equations. Furthermore, we propose a new distance
function related to the angle between two Plücker lines and the corresponding
Jacobian matrix to prevent violations of joint limits and avoid undesired
end-effector orientations. Differently from other approaches, such
as \cite{Dariush2010a}, our approach has a formal proof showing that
the second-order inequality constraints for collision avoidance always
prevent collisions between any pair of primitives.

\section{Task-Space Control Using Quadratic Programming}

A classic strategy used in differential inverse kinematic problems
consists in solving an optimization problem that minimizes the joint
velocities, $\dot{\myvec q}\in\mathbb{R}^{n}$, in the $l_{2}$-norm
sense. Given a desired task $\myvec x_{d}\in\mathbb{R}^{m}$, where
$\dot{\myvec x}_{d}=\myvec 0,\,\forall t$, and the task error $\tilde{\myvec x}\triangleq\myvec x-\myvec x_{d}$,
the control input $\myvec u$ is obtained as
\begin{gather}
\argmin{\dot{\myvec q}}{\norm{\mymatrix J\dot{\myvec q}+\eta\tilde{\myvec x}}_{2}^{2}+\lambda^{2}\norm{\dot{\myvec q}}_{2}^{2}}{\mymatrix W\dot{\myvec q}\leq\myvec w,}{\myvec u\in\,}\label{eq:minProblem_qp}
\end{gather}
where $\mymatrix J\in\mathbb{R}^{m\times n}$ is the task Jacobian,
${\lambda\in\left[0,\infty\right)}$ is a damping factor, and $\mymatrix W\in\mathbb{R}^{l\times n}$
and $\myvec w\in\mathbb{R}^{l}$ are used to impose linear constraints
in the control inputs.

An analogous scheme can be used to perform the minimization at the
joint acceleration level. Given the desired error dynamics $\ddot{\tilde{\myvec x}}+k_{d}\dot{\tilde{\myvec x}}+k_{p}\tilde{\myvec x}=\myvec 0,$
with ${k_{d},k_{p}\in\left(0,\infty\right)}$ such that ${k_{d}^{2}-4k_{p}>0}$
to obtain a non-oscillatory exponential error decay, the control input
$\myvec u_{a}$ is obtained as
\begin{gather}
\argmin{\ddot{\myvec q}}{\norm{\mymatrix J\ddot{\myvec q}+\myvec{\beta}}_{2}^{2}+\lambda^{2}\norm{\ddot{\myvec q}}_{2}^{2}}{\mathcal{\mymatrix{\Lambda}}\ddot{\myvec q}\leq\myvec{\zeta},}{\myvec u_{a}\in\,}\label{eq:min_Problem_qpp}
\end{gather}
where ${\myvec{\beta}=\left(k_{d}\mymatrix J+\dot{\mymatrix J}\right)\dot{\myvec q}+k_{p}\tilde{\myvec x}}$,
$\mathcal{\mymatrix{\Lambda}}\triangleq\begin{bmatrix}\mymatrix I & -\mymatrix I & \mymatrix W^{T}\end{bmatrix}^{T}$
and $\myvec{\zeta}\triangleq\left[\myvec{\gamma}_{u}^{T},-\myvec{\gamma}_{l}^{T},\myvec w^{T}\right]^{T}$,
with $\mathcal{\mymatrix{\Lambda}}\in\mathbb{R}^{\left(2n+l\right)\times n}$
and $\myvec{\zeta}\in\mathbb{R}^{2n+l}$. Analogously to (\ref{eq:minProblem_qp}),
$\mymatrix W$ and $\myvec w$ are used to impose arbitrary linear
constraints in the acceleration inputs, and two additional constraints,
that is $\myvec{\gamma}_{l}\leq\ddot{\myvec q}\leq\myvec{\gamma}_{u}$,
are imposed to minimize the joints velocities by limiting the joints
accelerations. More specifically, we define the acceleration lower
bound $\myvec{\gamma}_{l}$ and the acceleration upper bound $\myvec{\gamma}_{u}$,
respectively:
\begin{align}
\myvec{\gamma}_{l} & \triangleq k\left(-\mymatrix 1_{n}g\left(\dot{\tilde{\myvec x}}\right)-\dot{\myvec q}\right), & \myvec{\gamma}_{u} & \triangleq k\left(\mymatrix 1_{n}g\left(\dot{\tilde{\myvec x}}\right)-\dot{\myvec q}\right),\label{eq:acceleration_constraints}
\end{align}
where ${k\in\left[0,\infty\right)}$ is used to scale the feasible
region, $g:\mathbb{R}\to[0,\infty)$ is a positive definite nondecreasing
function (e.g., $g\left(\dot{\tilde{\myvec x}}\right)\triangleq\norm{\dot{\tilde{\myvec x}}}_{2}$)
and $\myvec 1_{n}$ is a $n$-dimensional column vector composed of
ones. As the bounds (\ref{eq:acceleration_constraints}) depend on
the error velocity $\dot{\tilde{\myvec x}}$, then $\myvec{\gamma}_{l}\to\myvec{\gamma}_{u}$
when $\dot{\tilde{\myvec x}}\to\myvec 0$. Therefore, $\myvec{\gamma}_{l}=\myvec{\gamma}_{u}=\myvec{\gamma}$
and $\myvec{\gamma}\leq\ddot{\myvec q}\leq\myvec{\gamma}$ becomes
$\ddot{\myvec q}=\myvec{\gamma}=-k\dot{\myvec q}$, whose solution
is given by $\dot{\myvec q}\left(t\right)=\dot{\myvec q}\left(0\right)\exp\left(-kt\right)$;
that is, as the task velocity goes to zero, the robot stops accordingly.\footnote{Those bounds are necessary because (\ref{eq:min_Problem_qpp}) minimizes
the joints \emph{accelerations}. Without them, the objective function
can be minimized even if the joints velocities are not null.}

If the robot is commanded by means of torque inputs (i.e., $\myvec{\tau}\triangleq\myvec u_{\tau}$),
we use (\ref{eq:min_Problem_qpp}) to compute $\myvec u_{a}$ and
the Euler-Langrange equation $\mymatrix M\ddot{\myvec q}+\myvec n=\myvec{\tau}$,
where $\mymatrix M\in\mathbb{R}^{n\times n}$ is the inertia matrix
and $\myvec n\in\mathbb{R}^{n}$ denotes the nonlinear terms including
Coriolis and gravity forces, to compute the control input
\begin{equation}
\myvec u_{\tau}=\myvec n+\mymatrix M\myvec u_{a}.\label{eq:Torques}
\end{equation}

\section{Vector Field Inequalities}

The VFI framework is composed of differential inequalities that are
used to avoid collisions between pairs of geometrical primitives \cite{Faverjon,Marinho2018}.
It requires a signed distance function $d\left(t\right)\in\mathbb{R}$
between two collidable objects and the Jacobian matrix $\mymatrix J_{d}$
relating the robot joint velocities with the time derivative of the
distance; that is, $\dot{d}\left(t\right)=\mymatrix J_{d}\dot{\myvec q}.$

In order to keep the robot outside a collision zone, the error distance
is defined as ${\tilde{d}\left(t\right)\triangleq d\left(t\right)-d_{\mathrm{safe}}}$,
where ${d_{\mathrm{safe}}\in\left[0,\infty\right)}$ is an arbitrary
constant safe distance, and the following inequalities must hold for
all $t$ \cite{Marinho2018}:
\begin{align}
\dot{\tilde{d}}\left(t\right)\geq- & \eta_{d}\tilde{d}\left(t\right)\Longleftrightarrow-\mymatrix J_{d}\dot{\myvec q}\leq\eta_{d}\tilde{d}\left(t\right),\label{eq:Constraint}
\end{align}
where $\eta_{d}\in[0,\infty)$ is used to adjust the approach velocity.
The lower is $\eta_{d}$, the lower is the approach velocity.

Alternatively, if we consider acceleration inputs, the VFI can be
extended by means of a second-order differential inequality \cite{Bouyarmane2018b}
\begin{align}
\ddot{\tilde{d}}\left(t\right) & \geq-\eta_{1}\dot{\tilde{d}}\left(t\right)-\eta_{2}\tilde{d}\left(t\right)\Longleftrightarrow-\mymatrix J_{d}\ddot{\myvec q}\leq\beta_{d},\label{eq:SecondInequality}
\end{align}
where $\beta_{d}=\left(\eta_{1}\mymatrix J_{d}+\dot{\mymatrix J}_{d}\right)\dot{\myvec q}+\eta_{2}\tilde{d}\left(t\right)$,
and $\eta_{1},\eta_{2}\in[0,\infty)$ are used to adjust the approach
acceleration. We enforce the condition ${\eta_{1}^{2}-4\eta_{2}>0}$
to obtain, in the worst case, a non-oscillatory exponential approach.

Analogously, if it is desired to keep the robot inside a safe region
(i.e., $\tilde{d}\left(t\right)\leq0$), the constraints (\ref{eq:Constraint})
and (\ref{eq:SecondInequality}) are rewritten, respectively, as
\begin{align}
\mymatrix J_{d}\dot{\myvec q} & \leq-\eta_{d}\tilde{d}\left(t\right), & \mymatrix J_{d}\ddot{\myvec q} & \leq-\beta_{d}.\label{eq:jointAccelerationInequality2}
\end{align}

In order to show that the inequality in (\ref{eq:SecondInequality})
prevents collisions with static obstacles---i.e., $\tilde{d}\left(t\right)\geq0$,
$\forall t\in\left[0,\infty\right)$---in robot commanded by means
of acceleration inputs,\footnote{Actual robots are usually commanded by means of velocity or torque
inputs. In the latter case, acceleration inputs are transformed into
torque inputs by using (\ref{eq:Torques}). Furthermore, we assume
that the actuators do not work in saturation in order to ensure feasibility
of (\ref{eq:SecondInequality}) and (\ref{eq:jointAccelerationInequality2}).} we propose the following lemma.
\begin{lem}
Consider the inequality constraint (\ref{eq:SecondInequality}), with
${\eta_{1},\eta_{2}\in(0,\infty)}$, ${\eta_{1}^{2}-4\eta_{2}>0}$,
${\tilde{d}\left(0\right)\triangleq\tilde{d}_{0}\in(0,\infty)}$,
${\dot{\tilde{d}}\left(0\right)\triangleq\dot{\tilde{d}}_{0}\in(-\infty,\infty)}$
and ${\eta_{1}\geq2\abs{\dot{\tilde{d}}_{0}}/\tilde{d}_{0}}$. Let
$\tilde{d}\left(t\right)$ be a smooth function for all $t\in\left[0,\infty\right)$,
then $\tilde{d}\left(t\right)>0$, $\forall t\in\left[0,\infty\right)$.
\end{lem}
\begin{IEEEproof}
First consider ${\ddot{\tilde{d}}\left(t\right)=-\eta_{1}\dot{\tilde{d}}\left(t\right)-\eta_{2}\tilde{d}\left(t\right)}$,
whose solution is given by ${\tilde{d}\left(t\right)=\left(r_{1}-r_{2}\right)^{-1}\left(c_{1}\exp\left(r_{1}t\right)+c_{2}\exp\left(r_{2}t\right)\right)},$
where ${c_{1}=\dot{\tilde{d}}_{0}-\tilde{d}_{0}r_{2}}$, ${c_{2}=\tilde{d}_{0}r_{1}-\dot{\tilde{d}}_{0}}$
and ${r_{1},r_{2}\in\left(-\infty,0\right)}$ are the roots of the
characteristic equation. Using the Comparison Lemma\footnote{Let $\dot{u}=f\left(t,u\right)$, $\dot{v}\leq f\left(t,v\right)$,
where $f$ is continuous in $t$ and in the functions $u(t)$ and
$v(t)$, and $v_{0}\leq u_{0}$; then $v(t)\leq u(t)\,\,\,\forall t\in\left[0,T\right)$.
Because it is possible to reduce any \emph{$n^{\text{th}}$} order
linear differential equation to $n$ first-order linear ones, the
Comparison Lemma, which is defined to first order systems, can be
extended to higher order linear systems in a straightforward way.} \cite{Khalil_ComparisonLemma}, it can be shown that the solution
of (\ref{eq:SecondInequality}) ${\forall t\in\left[0,\infty\right)}$
is ${d\left(t\right)\geq\left(r_{1}-r_{2}\right)^{-1}\left(c_{1}\exp\left(r_{1}t\right)+c_{2}\exp\left(r_{2}t\right)\right)}.$

Both $f_{1}\left(t\right)\triangleq\exp\left(r_{1}t\right)$ and $f_{2}\triangleq\exp\left(r_{2}t\right)$
are decreasing monotonic functions where $f_{2}\left(t\right)$ decreases
at a higher rate than $f_{1}\left(t\right)$ because $r_{1},r_{2}\in\left(-\infty,0\right)$
and ${r_{2}<r_{1}}$. Therefore, by inspection, ${\tilde{d}\left(t\right)\geq0,\quad\forall t\in\left[0,\infty\right)}$
if $c_{1}\geq0$ and $c_{1}\geq\abs{c_{2}}$, which is satisfied when
${\dot{\tilde{d}}_{0}\geq\tilde{d}_{0}r_{2}}$ and $\dot{\tilde{d}}_{0}\geq\tilde{d}_{0}\left(r_{1}+r_{2}\right)/2$.
Because $\dot{\tilde{d}}_{0}\geq\tilde{d}_{0}\left(r_{1}+r_{2}\right)/2\geq\tilde{d}_{0}r_{2}$
and $\eta_{1}=-\left(r_{1}+r_{2}\right)$, then $\eta_{1}\geq-2\dot{\tilde{d}}_{0}/\tilde{d}_{0}$.
Since $2\abs{\dot{\tilde{d}}_{0}}/\tilde{d}_{0}\geq-2\dot{\tilde{d}}_{0}/\tilde{d}_{0}$
then we choose $\eta_{1}\geq2\abs{\dot{\tilde{d}}_{0}}/\tilde{d}_{0}$,
which concludes the proof.
\end{IEEEproof}
Since we consider arbitrary values for $\dot{\tilde{d}}_{0}$, additional
bounds on the accelerations (i.e., $\ddot{\myvec q}_{\min}\leq\ddot{\myvec q}\leq\ddot{\myvec q}_{\max}$)
may result in the unfeasibility of constraints (\ref{eq:SecondInequality})
and (\ref{eq:jointAccelerationInequality2}). Therefore, existing
techniques \cite{DelPetre2018} may be adapted to determine the maximum
bounds on the accelerations such that the VFI can still be satisfied.

\section{Distance Functions and Jacobians}

Each VFI requires a distance function between two geometric primitives
and the Jacobian that relates the robot joint velocities to the time-derivative
of the distance function. Marinho et al. \cite{Marinho2018} presented
some useful distance functions and their corresponding Jacobians based
on pair of geometric primitives composed of Plücker lines, planes,
and points. 

In some applications, it is advantageous to define target regions,
instead of one specific position and/or orientation, in order to relax
the task and, therefore, release some of the robot degrees of freedom
(DOF) for additional tasks. For instance, if the task consists in
carrying a cup of water from one place to another, one way to perform
it is by defining a time-varying trajectory that determines the robot
end-effector pose at all times, which requires six DOF. An alternative
is to define a target position to where the cup of water should be
moved (which requires three DOF) while ensuring that the cup is not
too tilted to prevent spilling, which require one more DOF (i.e.,
the angle between a vertical line and the line passing through the
center of the cup, as shown in Fig.~\ref{fig.DistanceObjects-1-1}).
The task can be further relaxed by using only the distance to the
final position, which requires only one DOF, plus the angle constraint,
which requires, when activated, one more DOF.

The idea of using a conic constraint, which is equivalent to constraining
the angle between two intersecting lines, was first proposed by Gienger
et al. \cite{Gienger2006}. However, their description is singular
when the angle equals $k\pi$, for all $k\in\mathbb{Z}$. Therefore,
we propose a singularity-free conic constraint based on VFI. Since
this constraint requires a new Jacobian, namely the line-static-line
angle Jacobian, we briefly review the line Jacobian \cite{Marinho2018}.

\subsection{Line Jacobian}

Given a frame $\mathcal{F}_{z}$ attached to the robot kinematic chain,
let $\dq l_{z}$ be a dynamic Plücker line collinear to the $z$-axis
of $\mathcal{F}_{z}$. The line $\dq l_{z}$ and its time derivative
$\dot{\dq l}_{z}$ are described with respect to the inertial frame
$\frame{}$ as \cite{Marinho2018}
\begin{align}
\dq l_{z} & =\begin{bmatrix}\quat l_{z}\\
\quat m_{z}
\end{bmatrix}\in\mathbb{R}^{6}, & \dot{\dq l}_{z}= & \left[\begin{array}{c}
\mymatrix J_{l_{z}}\\
\mymatrix J_{m_{z}}
\end{array}\right]\dot{\myvec q}\triangleq\mymatrix J_{\dq l_{z}}\dot{\myvec q}.\label{PluckerLine}
\end{align}
where $\quat l_{z}\in\mathbb{R}^{3}$, with $\norm{\quat l_{z}}_{2}=1$,
is the line direction, and $\quat p_{z}\times\quat l_{z}=\quat m_{z}\in\mathbb{R}^{3}$
is the line moment, with $\quat p_{z}\in\mathbb{R}^{3}$ being an
arbitrary point on the line.

\subsection{Line-static-line-angle Jacobian, $\protect\mymatrix J_{\phi_{\protect\dq l_{z},\protect\dq l}}$}

Since the angle between $\dq l_{z}$ and the static Plücker line $\dq l=\begin{bmatrix}\myvec l^{T} & \myvec m^{T}\end{bmatrix}^{T}$
(i.e., $\dot{\dq l}=\myvec 0$, $\forall t$), which is given by $\phi_{\dq l_{z},\dq l}=\arccos\left(\dotproduct{\quat l_{z},\quat l}\right)$,
has a singular time derivative when $\dotproduct{\quat l_{z},\quat l}=\pm1$,
the function $f:\left[0,\pi\right]\to\left[0,4\right]$ is proposed
instead:
\begin{align}
f\left(\phi_{\dq l_{z},\dq l}\right) & \triangleq\norm{\quat l_{z}-\quat l}_{2}^{2}=2-2\cos\phi_{\dq l_{z},\dq l}=\left(\quat l_{z}-\quat l\right)^{T}\left(\quat l_{z}-\quat l\right).\label{eq:f1}
\end{align}

As $f\left(\phi_{\dq l_{z},\dq l}\right)$ is a continuous bijective
function, controlling the distance function (\ref{eq:f1}) is equivalent
to controlling the angle $\phi_{\dq l_{z},\dq l}\in\left[0,\pi\right]$.

The time derivative of (\ref{eq:f1}) is given by
\begin{align}
\frac{d}{dt}f\left(\phi_{\dq l_{z},\dq l}\right) & =2\left(\quat l_{z}-\quat l\right)^{T}\frac{d}{dt}\left(\quat l_{z}-\quat l\right)=\underset{\mymatrix J_{\phi_{\dq l_{z},\dq l}}}{\underbrace{2\left(\quat l_{z}-\quat l\right)^{T}\mymatrix J_{l_{z}}}}\dot{\myvec q}.
\end{align}

\section{(Self)-Collision Avoidance Constraints\label{sec:(Self)-Collision-Avoidance-Const}}

\begin{figure}
\noindent \begin{centering}
\def\svgwidth{3\columnwidth} 
\vspace{-2mm}{\Huge{}\resizebox{1\columnwidth}{!}{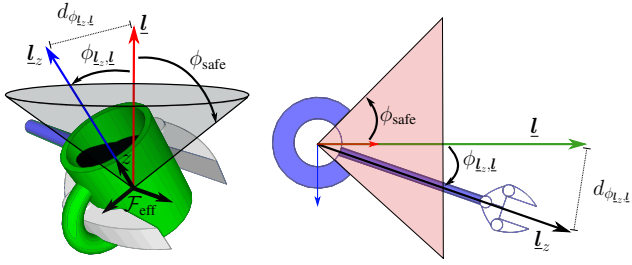}}{\Huge\par}
\par\end{centering}
\caption{Applications of the line-static-line-angle Jacobian. On the \emph{left},
the constraint is used to keep a bounded inclination of the end-effector
with respect to a vertical line passing through the origin of the
end-effector frame. On the \emph{right}, the constraint is used to
impose joint limits. The angle $\phi_{\protect\dq l_{z},\protect\dq l}$
between two Plücker lines is related to the distance $d_{\phi_{\protect\dq l_{z},\protect\dq l}}=f(\phi_{\protect\dq l_{z},\protect\dq l})^{1/2}$
by means of the law of cosines.\label{fig.DistanceObjects-1-1}}
\end{figure}

As shown in Fig.~\ref{fig.DistanceObjects-1-1}, the line-static-line-angle
constraint is used to prevent violation of joint limits (\emph{right})
and also to avoid undesired end-effector orientations (\emph{left}).
In order to prevent violation of joint limits, for each joint we place
a line, $\dq l$, perpendicular to the joint rotation axis and in
the middle of its angular displacement range. We also place a line,
$\dq l_{z}$, along the link attached to the joint; therefore, the
angle distance between them is calculated by using (\ref{eq:f1}).
Since the angle distance is constrained between $\left[0,\pi\right]$
in any direction (the other one is equivalent to $\left[-\pi,0\right]$),
the maximum joint range limit can be defined in the interval $\left[0,2\pi\right]$.
This strategy requires one constraint per joint instead of two, as
in other approaches \cite{Conference1994,Guo2014a}, because those
approaches need to define constraints for both upper and lower joint
limits.

Likewise, this strategy allows defining a maximum end-effector angle
to avoid undesired orientations. In this case, however, the angle
is obtained between a vertical line, $\dq l$, passing through the
origin of the end-effector frame and a line collinear with the $z$-axis
of the end-effector frame, $\dq l_{z}$. Given a maximum angle $\phi_{\text{safe}}$,
the constraint $\phi_{\dq l_{z},\dq l}\leq\phi_{\text{safe}}$ defines
a cone whose centerline is given by $\dq l$.

The corresponding first-order and second-order VFIs used to constrain
the angle inside a safe cone are given, respectively, by
\begin{align}
\mymatrix J_{\phi_{\dq l_{z},\dq l}}\dot{\myvec q} & \leq-\eta\tilde{f}\left(\phi_{\dq l_{z},\dq l}\right),\label{eq:constraint_phi}\\
\mymatrix J_{\phi_{\dq l_{z},\dq l}}\ddot{\myvec q} & \leq-\beta_{\phi_{\dq l_{z},\dq l}},\label{eq:constrant_phi_qpp}
\end{align}
where $\tilde{f}\left(\phi_{\dq l_{z},\dq l}\right)\triangleq f\left(\phi_{\dq l_{z},\dq l}\right)-f\left(\phi_{\text{safe}}\right)$
and $\beta_{\phi_{\dq l_{z},\dq l}}=\left(\eta_{1}\mymatrix J_{\phi_{\dq l_{z},\dq l}}+\dot{\mymatrix J}_{\phi_{\dq l_{z},\dq l}}\right)\dot{\myvec q}+\eta_{2}\tilde{f}\left(\phi_{\dq l_{z},\dq l}\right)$.

\begin{figure}
\def\svgwidth{3\columnwidth} 
\vspace{-2mm} 
\noindent \begin{centering}
{\Huge{}\resizebox{1\columnwidth}{!}{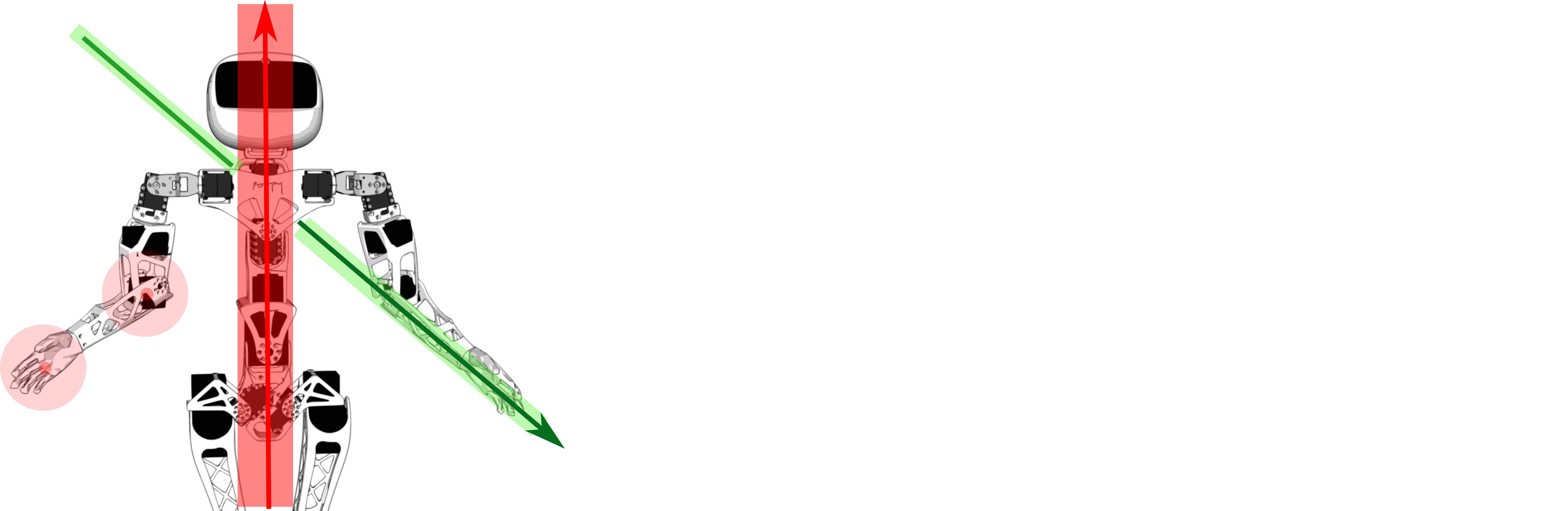}}{\Huge\par}
\par\end{centering}
\caption{On the \emph{left}, the tobot description using geometric primitives.
The torso and the forearms are modeled with infinite cylinders and
spheres. On the \emph{middle}, a target region is defined to the right
hand. On the \emph{right}, planes are used to constrain the right
hand into an admissible space towards the target region.\label{fig.PoppyConstraints}\label{fig.DistanceObjects-1}}

\vspace{-3mm}
\end{figure}

In order to prevent self-collisions, we modeled the robot with spheres
and cylinders, as shown in Fig.~\ref{fig.PoppyConstraints}.

The line-static-line constraint \cite{Marinho2018} is used to prevent
collisions between the arm and the torso, where the line $\dq l_{z}$
is located along the torso and the line $\dq l$ is placed along the
forearm. In case that the distance $d_{\dq l_{z},\dq l}$ between
the lines is zero but there is no collision, which could happen because
lines are infinite, the constraint is disabled and a point-static-line
constraint is used instead. In this case, the point located at the
hand or the one located at the elbow is used, depending which one
is closest to the line. The torso-arm constraint is written as
\begin{align}
-\mymatrix J_{\mathrm{tarm}}\dot{\myvec q} & \leq\eta\tilde{d}_{\mathrm{tarm}},\label{eq:Constraint_Tarm}
\end{align}
where $\mymatrix J_{\mathrm{tarm}}=\mymatrix J_{\dq l_{z},\dq l}$
and $\tilde{d}_{\mathrm{tarm}}=d_{\dq l_{z},\dq l}-d_{\mathrm{safe}}$,
if $d_{\dq l_{z},\dq l}\neq0$, or $\mymatrix J_{\mathrm{tarm}}=\mymatrix J_{p,\dq l}$
and $\tilde{d}_{\mathrm{tarm}}=d_{p,\dq l}-d_{\mathrm{safe}}$ otherwise,
where $d_{p,\dq l}$ is the point-static-line distance, $\mymatrix J_{p,\dq l}$
and $\mymatrix J_{\dq l_{z},\dq l}$ are the point-static-line and
the line-static-line Jacobians, respectively \cite{Marinho2018}.

The inequality constraints in terms of the joints accelerations are
written in similar way, as shown in (\ref{eq:constrant_phi_qpp}),
and therefore they will be omitted for the sake of conciseness.

To prevent collisions with obstacles, which are modeled with spheres,
cylinders or planes, we use additional constraints. Fig.~\ref{fig.DistanceObjects-1}
shows an application based on the plane constraint, where four lateral
planes are used to constrain the right hand of the robot inside a
desired region. This way, it is impossible for the robot hand to reach
the target region from below because the hand cannot trespass any
of the four lateral planes. These four additional constraints are
written as
\begin{align}
-\mymatrix J_{p,n_{\dq{\pi}_{i}}}\dot{\myvec q} & \leq\eta\tilde{d}_{p,n_{\dq{\pi}_{i}}},\label{eq:constraint_planePoint}
\end{align}
where $i=\{1,2,3,4\}$ and $\tilde{d}_{p,n_{\dq{\pi}_{i}}}=d_{p,n_{\dq{\pi}_{i}}}-d_{\pi,\mathrm{safe}}$,
with $d_{\pi,\mathrm{safe}}$ being the safe distance to each plane
and $d_{p,n_{\dq{\pi}_{i}}}$ and $\mymatrix J_{p,n_{\dq{\pi}_{i}}}$
are the point-static-plane distance and its Jacobian, respectively
\cite{Marinho2018}.

\section{Simulations and Experimental Results}

To validate the proposed approach, we performed simulations on V-REP\footnote{http://www.coppeliarobotics.com/}
and Matlab\footnote{https://www.mathworks.com/}, and real experiments
using ROS\footnote{https://www.ros.org/} and Python. We used DQ Robotics\footnote{https://dqrobotics.github.io/}
for both robot modeling and the description of the geometric primitives.

We implemented two controllers based on quadratic programming, namely
QP and CQP, where QP does not take into account constraints and CQP
denotes the constrained case, and validated them in both simulation
and in a real-time implementation on a real robot.

\subsection{Simulation setup\label{subsec:Simulation-setup}}

The goal is to control the left-hand of a humanoid robot to put a
cup on a table while keeping the cup tilting below threshold (constraint~(\ref{eq:constraint_phi})),
avoiding reaching joint limits (one constraint~(\ref{eq:constraint_phi})
for each joint) and preventing self-collisions (constraint~(\ref{eq:Constraint_Tarm}))
and collisions with the workspace (constraint~(\ref{eq:constraint_planePoint})),
as shown in Fig.~\ref{fig:Control-of-theSetup-2_experiment}. To
accomplish this goal, the control law minimizes the distance between
the left-hand and the table, which is modeled as a plane, and the
control inputs take into account the kinematic chain composed of the
5-DOF torso and the 4-DOF left arm, which has a total of nine DOF.
The safe angle, $\phi_{\mathrm{safe}}=0.1\unit{rad}$, denotes the
cup maximum tilting angle, as shown in Fig.~\ref{fig.DistanceObjects-1-1}.
We defined four additional planes, as shown in Fig.~\ref{fig.DistanceObjects-1},
to prevent reaching the table from below, where $d_{\pi,\mathrm{safe}}=0.01$
for all four planes. Furthermore, the safe distance between the arm
and the torso is $d_{\mathrm{safe}}\triangleq d_{\textrm{tarm}}=0.07$.
The same setup was used for the simulation using velocity inputs and
the one using torque inputs.

\subsection{Task space control: joint velocity inputs\label{subsec:Kinematic-Control}}

In this simulation, for both QP and CQP, the objective function is
defined as in (\ref{eq:minProblem_qp}), where $\tilde{\myvec x}\triangleq d_{p,n_{\dq{\pi}}}$
is the distance between the end-effector and the table, and $\mymatrix J\triangleq\mymatrix J_{p,n_{\dq{\pi}}}$
is the corresponding point-static-plane Jacobian matrix. Furthermore,
the convergence parameter ($\eta=0.36$) and the damping factor ($\lambda=0.01$)
were defined empirically in order to provide a smooth and fast enough
convergence rate.

In the constrained case (CQP), the control inputs $\myvec u\triangleq\dot{\myvec q}$
are computed using (\ref{eq:minProblem_qp}) with the aforementiond
task-Jacobian and error function, where ${\mymatrix W=\begin{bmatrix}\mymatrix J_{\phi_{\dq l_{z},\dq l}}^{T} & -\mymatrix J_{p,n_{\dq{\pi}_{\mathcal{I}}}}^{T} & \mymatrix J_{\phi_{\dq l_{z\mathcal{I}},\dq l_{\mathcal{I}}}}^{T} & -\mymatrix J_{\mathrm{tarm}}^{T}\end{bmatrix}^{T}}$
with
\begin{align}
\mymatrix J_{p,n_{\dq{\pi}_{\mathcal{I}}}} & =\begin{bmatrix}\mymatrix J_{p,n_{\dq{\pi}_{1}}}^{T} & \cdots & \mymatrix J_{p,n_{\dq{\pi}_{4}}}^{T}\end{bmatrix}^{T},\label{eq:plane_jacobians}\\
\mymatrix J_{\phi_{\dq l_{z\mathcal{I}},\dq l_{\mathcal{I}}}} & =\begin{bmatrix}\mymatrix J_{\phi_{\dq l_{z_{1}},\dq l_{1}}}^{T} & \cdots & \mymatrix J_{\phi_{\dq l_{z_{9}},\dq l_{9}}}^{T}\end{bmatrix}^{T},\label{eq:angle_jacobians}
\end{align}
and $\myvec w=\eta\begin{bmatrix}-\tilde{f}\left(\phi_{\dq l_{z},\dq l}\right) & \tilde{\myvec d}_{p,n_{\dq{\pi}_{\mathcal{I}}}}^{T} & -\tilde{\myvec F}^{T} & \tilde{d}_{\mathrm{tarm}}\end{bmatrix}^{T}$
with 
\begin{align}
\tilde{\myvec d}_{p,n_{\dq{\pi}_{\mathcal{I}}}} & =\begin{bmatrix}\tilde{d}_{p,n_{\dq{\pi}_{1}}} & \cdots & \tilde{d}_{p,n_{\dq{\pi}_{4}}}\end{bmatrix}^{T},\label{eq:plane_distances}\\
\tilde{\myvec F} & =\begin{bmatrix}\tilde{f}\left(\phi_{\dq l_{z_{1}},\dq l_{1}}\right) & \cdots & \tilde{f}\left(\phi_{\dq l_{z_{9}},\dq l_{9}}\right)\end{bmatrix}^{T}.\label{eq:angle_distances}
\end{align}
Eq.~(\ref{eq:plane_jacobians}) and (\ref{eq:plane_distances}) are
used to enforce the four plane constraints (recall (\ref{eq:constraint_planePoint})),
whereas (\ref{eq:angle_jacobians}) and (\ref{eq:angle_distances})
are used to avoid joint limits in all nine joints.

\begin{figure}
\def\svgwidth{1.6\columnwidth}
\vspace{-2mm}
\noindent \begin{centering}
{\large{}\resizebox{1\columnwidth}{!}{
\begingroup%
  \makeatletter%
  \providecommand\color[2][]{%
    \errmessage{(Inkscape) Color is used for the text in Inkscape, but the package 'color.sty' is not loaded}%
    \renewcommand\color[2][]{}%
  }%
  \providecommand\transparent[1]{%
    \errmessage{(Inkscape) Transparency is used (non-zero) for the text in Inkscape, but the package 'transparent.sty' is not loaded}%
    \renewcommand\transparent[1]{}%
  }%
  \providecommand\rotatebox[2]{#2}%
  \newcommand*\fsize{\dimexpr\f@size pt\relax}%
  \newcommand*\lineheight[1]{\fontsize{\fsize}{#1\fsize}\selectfont}%
  \ifx\svgwidth\undefined%
    \setlength{\unitlength}{406.28855896bp}%
    \ifx\svgscale\undefined%
      \relax%
    \else%
      \setlength{\unitlength}{\unitlength * \real{\svgscale}}%
    \fi%
  \else%
    \setlength{\unitlength}{\svgwidth}%
  \fi%
  \global\let\svgwidth\undefined%
  \global\let\svgscale\undefined%
  \makeatother%
  \begin{picture}(1,0.26805803)%
    \lineheight{1}%
    \setlength\tabcolsep{0pt}%
    \put(0,0){\includegraphics[width=\unitlength,page=1]{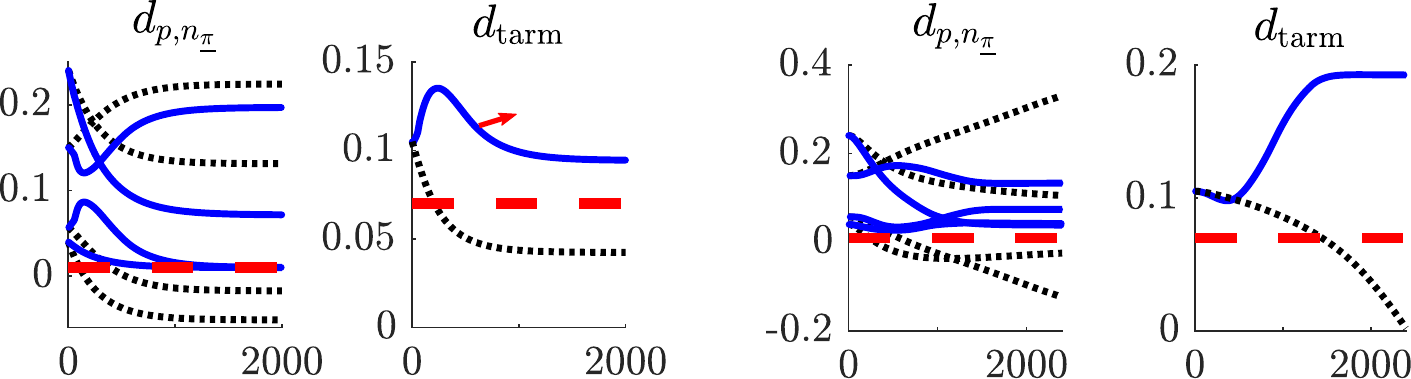}}%
    \put(0.37325921,0.18317304){\color[rgb]{0,0,0}\makebox(0,0)[lt]{\lineheight{1.25}\smash{\begin{tabular}[t]{l}$\textrm{CQP}$\end{tabular}}}}%
    \put(0,0){\includegraphics[width=\unitlength,page=2]{images/dplanes2.pdf}}%
    \put(0.3486373,0.05374497){\color[rgb]{0,0,0}\makebox(0,0)[lt]{\lineheight{1.25}\smash{\begin{tabular}[t]{l}$\textrm{QP}$\end{tabular}}}}%
    \put(0,0){\includegraphics[width=\unitlength,page=3]{images/dplanes2.pdf}}%
    \put(0.86281637,0.04469499){\color[rgb]{0,0,0}\makebox(0,0)[lt]{\lineheight{1.25}\smash{\begin{tabular}[t]{l}$\textrm{Lim}$\end{tabular}}}}%
  \end{picture}%
\endgroup%
}}{\large\par}
\par\end{centering}
\caption{Control of the left-hand in simulation. On the \emph{left}, the robot
kinematics is used with first-order VFI. On the \emph{right}, the
robot dynamics is used with second-order VFI.\label{fig:dplanes}}

\vspace{-3mm}
\end{figure}

Figure~\ref{fig:Control-of-theSetup-2_experiment}\emph{A} shows
the simulation snapshots. Although the task is fulfilled for both
controllers, only CQP prevents undesired cup orientations and respects
all other constraints, as shown in Fig.~\ref{fig:dplanes}.

\subsection{Task space control: joint torque inputs}

We performed a second simulation using the robot dynamic model with
second-order VFI. V-REP was used only as a visualization and collision-checking
tool. We assume perfect knowledge of the inertial parameters, which
are obtained directly from V-REP. The minimization is performed in
the joint accelerations using formulation (\ref{eq:min_Problem_qpp}),
where $\mymatrix J$ and $\tilde{\myvec x}$, which are needed to
obtain $\myvec{\beta}$, are the same as in Section~\ref{subsec:Kinematic-Control}.
In addition, all constraints are rewritten as in (\ref{eq:SecondInequality})
and (\ref{eq:jointAccelerationInequality2}) to obtain $\mymatrix W$
and $\myvec w$, where $\mymatrix W$ is exactly the same as in the
first-order case and $\myvec w=\left[\begin{array}{cccc}
-\beta_{\phi_{\dq l_{z},\dq l}} & \myvec{\beta}_{p,n_{\dq{\pi}_{\mathcal{I}}}}^{T} & -\myvec{\beta}_{\text{joints}}^{T} & \beta_{\mathrm{tarm}}\end{array}\right]^{T}$, with $\myvec{\beta}_{p,n_{\dq{\pi}_{\mathcal{I}}}}=\begin{bmatrix}\beta_{p,n_{\dq{\pi}_{1}}} & \cdots & \beta_{p,n_{\dq{\pi}_{4}}}\end{bmatrix}^{T}$
and $\myvec{\beta}_{\text{joints}}=\begin{bmatrix}\beta_{\phi_{\dq l_{z_{1}},\dq l_{1}}} & \cdots & \beta_{\phi_{\dq l_{z_{9}},\dq l_{9}}}\end{bmatrix}^{T}$.
Furthermore, $k_{d}=\eta_{1}=1.5$, $k_{p}=\eta_{2}=0.3$, and $g\left(\dot{\tilde{\myvec x}}\right)\triangleq\norm{\dot{\tilde{\myvec x}}}_{2}$.
Finally, the control inputs $\myvec u\triangleq\myvec{\tau}$ are
computed using (\ref{eq:Torques}).

\begin{figure}
\def\svgwidth{1.3\columnwidth}
\vspace{-2mm}
\noindent \begin{centering}
\resizebox{1\columnwidth}{!}{
\begingroup%
  \makeatletter%
  \providecommand\color[2][]{%
    \errmessage{(Inkscape) Color is used for the text in Inkscape, but the package 'color.sty' is not loaded}%
    \renewcommand\color[2][]{}%
  }%
  \providecommand\transparent[1]{%
    \errmessage{(Inkscape) Transparency is used (non-zero) for the text in Inkscape, but the package 'transparent.sty' is not loaded}%
    \renewcommand\transparent[1]{}%
  }%
  \providecommand\rotatebox[2]{#2}%
  \newcommand*\fsize{\dimexpr\f@size pt\relax}%
  \newcommand*\lineheight[1]{\fontsize{\fsize}{#1\fsize}\selectfont}%
  \ifx\svgwidth\undefined%
    \setlength{\unitlength}{381.08956146bp}%
    \ifx\svgscale\undefined%
      \relax%
    \else%
      \setlength{\unitlength}{\unitlength * \real{\svgscale}}%
    \fi%
  \else%
    \setlength{\unitlength}{\svgwidth}%
  \fi%
  \global\let\svgwidth\undefined%
  \global\let\svgscale\undefined%
  \makeatother%
  \begin{picture}(1,0.30785925)%
    \lineheight{1}%
    \setlength\tabcolsep{0pt}%
    \put(0,0){\includegraphics[width=\unitlength,page=1]{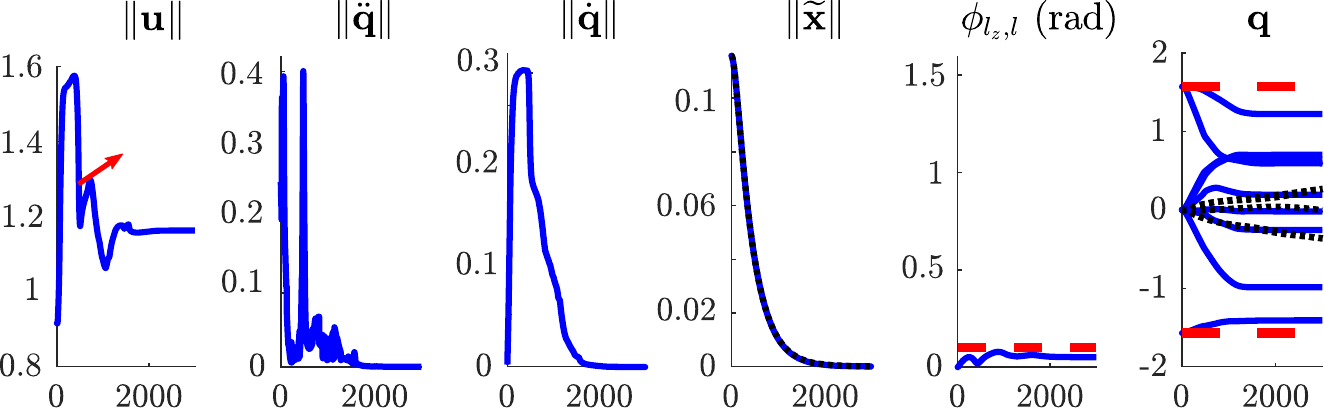}}%
    \put(0.0915537,0.19385779){\color[rgb]{0,0,0}\makebox(0,0)[lt]{\lineheight{1.25}\smash{\begin{tabular}[t]{l}$\textrm{CQP}$\end{tabular}}}}%
    \put(0,0){\includegraphics[width=\unitlength,page=2]{images/DynResults3.pdf}}%
    \put(0.1047598,0.1044249){\color[rgb]{0,0,0}\makebox(0,0)[lt]{\lineheight{1.25}\smash{\begin{tabular}[t]{l}$\textrm{QP}$\end{tabular}}}}%
    \put(0,0){\includegraphics[width=\unitlength,page=3]{images/DynResults3.pdf}}%
    \put(0.75010932,0.07976422){\color[rgb]{0,0,0}\makebox(0,0)[lt]{\lineheight{1.25}\smash{\begin{tabular}[t]{l}$\textrm{Lim}$\end{tabular}}}}%
    \put(0,0){\includegraphics[width=\unitlength,page=4]{images/DynResults3.pdf}}%
  \end{picture}%
\endgroup%
}
\par\end{centering}
\caption{Control of the left-hand using the robot dynamics and second-order
VFI in simulation. The \emph{dashed-black} line corresponds to QP
and the \emph{solid-blue line} denotes the CQP. From \emph{left} to
\emph{right}: norm of the control inputs (torques); norm of the joint
accelerations; norm of the joint velocities; norm of the task error;
angle $\phi_{\protect\dq l_{z},\protect\dq l}$; and the joints positions
of both torso and left arm. The \emph{red-dashed line} in the last
two graphs correspond to the maximum allowable tilting angle $\phi_{\text{safe}}$
and the joints limits, respectively. The horizontal axis corresponds
to iterations.\label{fig:dyn_results}}
\end{figure}

Fig.~\ref{fig:dyn_results} shows that both QP and CQP minimize the
task error and\emph{ }the joint accelerations, as expected. The joint
accelerations in CQP, however, are more abrupt because the collision-avoidance
constraints enforce abrupt changes in the robot velocities to prevent
collisions. Because of that, CQP required higher control effort, as
shown by larger values of $\norm{\myvec u}$ for CQP in Fig.~\ref{fig:dyn_results}.

Since CQP enforces the constraints $\myvec{\gamma}_{l}\leq\ddot{\myvec q}\leq\myvec{\gamma}_{u}$,
where $\myvec{\gamma}_{l}$ and $\myvec{\gamma}_{u}$ are calculated
according to (\ref{eq:acceleration_constraints}), the joint velocities
go to zero as the end-effector stops. As QP\emph{ }does not enforce
those constraints, the robot joints continue to move after the end-effector
stops, as shown in Fig.~\ref{fig:dyn_results}.

Finally, the cup maximum tilting angle is respected, as well as the
other constraints for collision avoidance and joint limit avoidance,
only when CQP is used, as shown in Fig.~\ref{fig:dplanes}.

\subsection{Experimental Results}

We performed a real experiment using the platform composed of the
upper body of the Poppy humanoid robot.\footnote{https://www.poppy-project.org/en/robots/poppy-humanoid}
We used Python and ROS Melodic in a computer running Ubuntu 18.04
64 bits, equipped with a Intel i7 4712HQ with 16GB RAM, in addition
to quadprog\footnote{https://pypi.org/project/quadprog/} to solve
(\ref{eq:minProblem_qp}). The solver required about $191\mu\textrm{s}\pm47\mu\textrm{s}$
to generate the control inputs in the constrained case. Furthermore,
it was required about $5.2\unit{ms}$$\pm$$1\unit{ms}$ to compute
15 constraint Jacobians (one angle constraint, four plane constraints,
one torso-left-arm constraint, and nine joint limit constraints).
However, we set the loop rate control to 50Hz due to low-level drivers
limitations.

\begin{figure}
\def\svgwidth{1.6\columnwidth}

\noindent \begin{centering}
{\large{}\resizebox{1\columnwidth}{!}{
\begingroup%
  \makeatletter%
  \providecommand\color[2][]{%
    \errmessage{(Inkscape) Color is used for the text in Inkscape, but the package 'color.sty' is not loaded}%
    \renewcommand\color[2][]{}%
  }%
  \providecommand\transparent[1]{%
    \errmessage{(Inkscape) Transparency is used (non-zero) for the text in Inkscape, but the package 'transparent.sty' is not loaded}%
    \renewcommand\transparent[1]{}%
  }%
  \providecommand\rotatebox[2]{#2}%
  \newcommand*\fsize{\dimexpr\f@size pt\relax}%
  \newcommand*\lineheight[1]{\fontsize{\fsize}{#1\fsize}\selectfont}%
  \ifx\svgwidth\undefined%
    \setlength{\unitlength}{385.87623596bp}%
    \ifx\svgscale\undefined%
      \relax%
    \else%
      \setlength{\unitlength}{\unitlength * \real{\svgscale}}%
    \fi%
  \else%
    \setlength{\unitlength}{\svgwidth}%
  \fi%
  \global\let\svgwidth\undefined%
  \global\let\svgscale\undefined%
  \makeatother%
  \begin{picture}(1,0.30732986)%
    \lineheight{1}%
    \setlength\tabcolsep{0pt}%
    \put(0,0){\includegraphics[width=\unitlength,page=1]{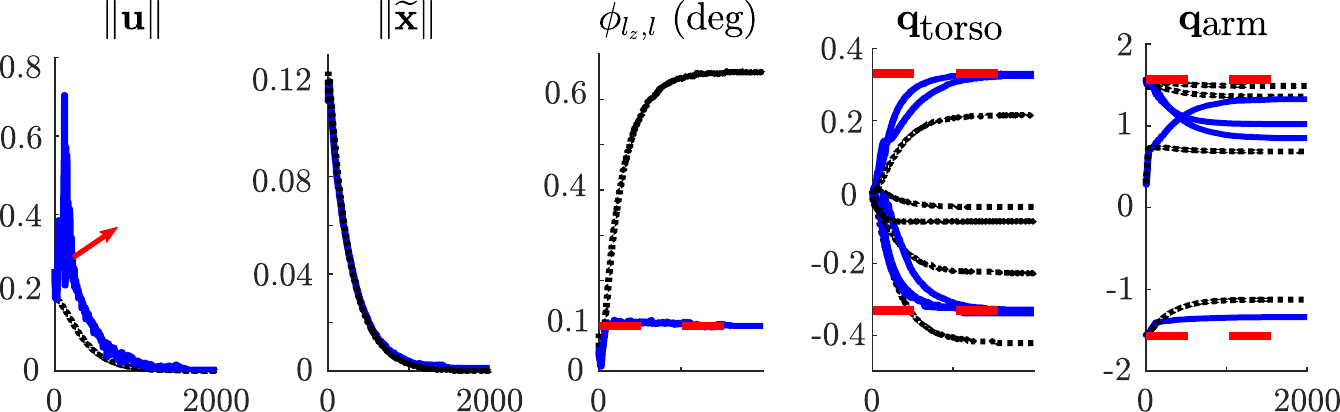}}%
    \put(0.0864485,0.13995426){\color[rgb]{0,0,0}\makebox(0,0)[lt]{\lineheight{1.25}\smash{\begin{tabular}[t]{l}$\textrm{CQP}$\end{tabular}}}}%
    \put(0,0){\includegraphics[width=\unitlength,page=2]{images/KinResults22new2.pdf}}%
    \put(0.09770241,0.09121117){\color[rgb]{0,0,0}\makebox(0,0)[lt]{\lineheight{1.25}\smash{\begin{tabular}[t]{l}$\textrm{QP}$\end{tabular}}}}%
    \put(0,0){\includegraphics[width=\unitlength,page=3]{images/KinResults22new2.pdf}}%
    \put(0.47471186,0.09991156){\color[rgb]{0,0,0}\makebox(0,0)[lt]{\lineheight{1.25}\smash{\begin{tabular}[t]{l}$\textrm{Lim}$\end{tabular}}}}%
    \put(0,0){\includegraphics[width=\unitlength,page=4]{images/KinResults22new2.pdf}}%
  \end{picture}%
\endgroup%
}}{\large\par}
\par\end{centering}
\caption{Control of the left-hand using the robot kinematics and first-order
VFI using the real-time implementation on the real robot. From \emph{left}
to \emph{right}: norm of the control inputs (joint velocities); norm
of the task error; the angle $\phi_{\protect\dq l_{z},\protect\dq l}$
between the cup centerline and a vertical line; joints configurations
of the torso; and the joints configurations of the left arm. The \emph{dashed
red line} denotes the maximum allowable tilting angle and the joint
limits. The horizontal axis corresponds to iterations.\label{fig:Control-of-theAngleError-1}}
\end{figure}

We used the same task specification and parameters defined in Sections~\ref{subsec:Simulation-setup}
and \ref{subsec:Kinematic-Control}. All collision-avoidance constraints
were activated, for both QP and CQP, to ensure the robot safety. This
way, the only difference between them is that QP does not have the
cup tilting angle constraint whereas CQP has. Fig.~\ref{fig:Control-of-theAngleError-1}
shows the task error decay and the angle $\phi_{\dq l_{z},\dq l}$
for both controllers. In both cases, the task is fulfilled with the
same convergence rate, which is determined by $\eta$. However, only
CQP respects the angle constraint and the limit joints constraints,
as expected.

\begin{figure}
\def\svgwidth{2\columnwidth}
\noindent \begin{raggedright}
\resizebox{1\columnwidth}{!}{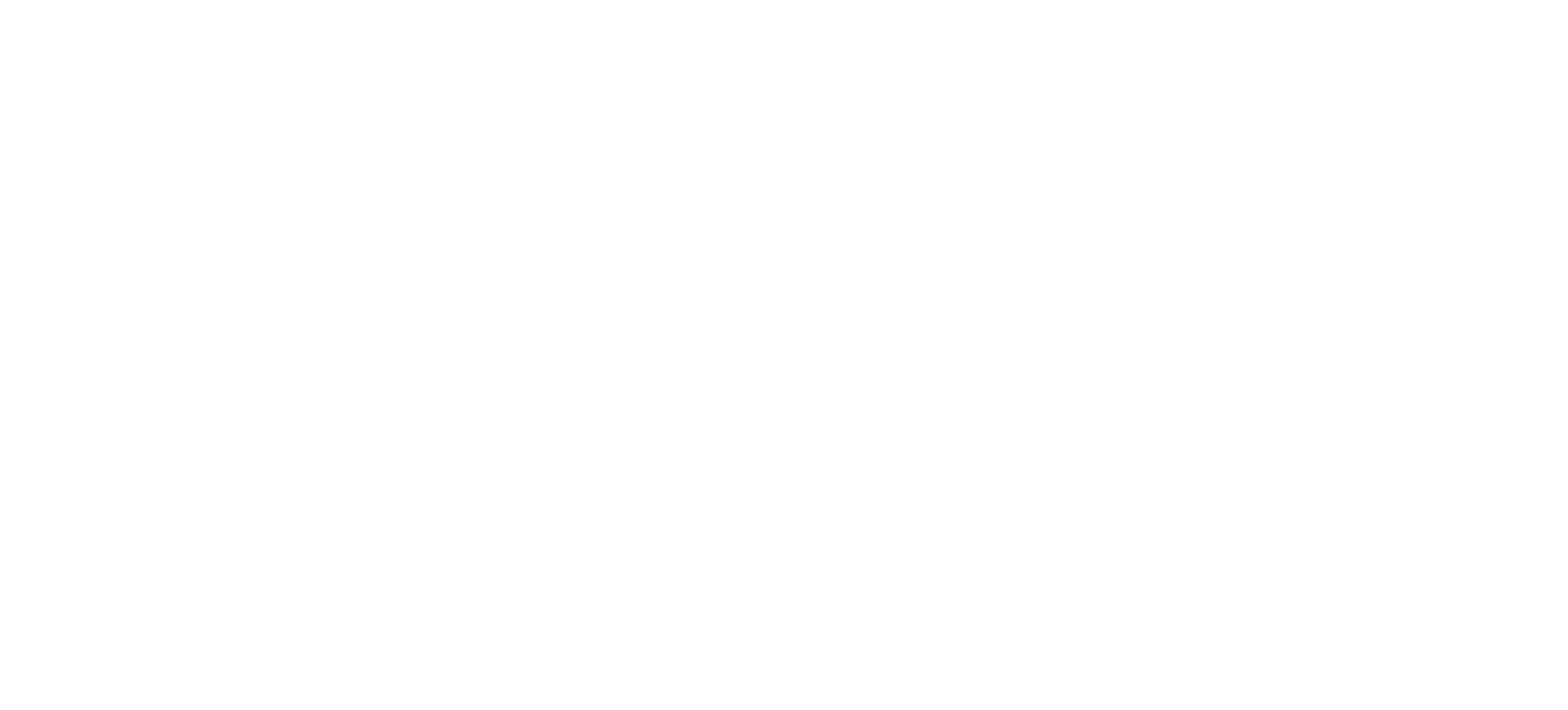}
\par\end{raggedright}
\caption{Control of the left-hand using the robot kinematics and first-order
VFI. \textbf{\emph{A}} (simulation): on \emph{top,} only the plane
constraints were used, whereas at the \emph{bottom} all constraints
were used. The\emph{ blue axis} denotes the cup orientation and the
\emph{red cone} represents the maximum allowable tilting for the \emph{blue
axis}. \emph{Red-shaded} body parts indicate a collision. \label{fig:Control-of-theSetup}\textbf{\emph{B}}
(real-time implementation on the real robot): on the \emph{top} three
snapshots\emph{,} the cup tilting angle constraint is disabled, whereas
on the three snapshots at the bottom this constraint is enabled. All
other collision-avoidance constraints are activated to ensure the
robot safety. The red square denotes the target region. \label{fig:Control-of-theSetup-2_experiment}}

\vspace{-5mm}
\end{figure}

Figure~\ref{fig:Control-of-theSetup-2_experiment}\emph{B} shows
that, in both cases, the end-effector reached the target zone, but
only CQP does not violate the cup tilting angle constraint, as expected.

\section{Conclusions}

This work extended the VFI framework, which was initially proposed
using first-order kinematics, to second-order kinematics. This allows
its use in applications that requires the robot dynamics by means
of the relationship between the joint torques and joint accelerations
in the Euler-Lagrange equations. Furthermore, we proposed a novel
singularity-free conic constraint to limit the angle between two Plücker
lines. This new constraint is used to prevent the violation of joint
limits and/or to avoid undesired end-effector orientations.

We evaluated the proposed method, using kinematic and dynamic formulations,
both in simulation and on a real humanoid robot. The results showed
that the robot always avoids collisions with static obstacles, self-collisions,
and undesired orientations while performing manipulation tasks. Our
approach requires an environment modeled with sufficient geometric
primitives and is not free of local minima, but it does provide reactive
behaviors. Future works will be focused on the implementation of the
VFI framework on a full humanoid to enforce balance constraints and
will also consider dynamic obstacles.

\bibliographystyle{IEEEtran}
\bibliography{IEEEexample,library}

\end{document}